\documentclass{article}

    \PassOptionsToPackage{numbers, sort&compress}{natbib}
\usepackage[preprint]{neurips_2023}

\usepackage[utf8]{inputenc} % allow utf-8 input
\usepackage[T1]{fontenc}    % use 8-bit T1 fonts
\usepackage[colorlinks]{hyperref}       % hyperlinks
\usepackage{url}            % simple URL typesetting
\usepackage{booktabs}       % professional-quality tables
\usepackage{amsfonts}       % blackboard math symbols
\usepackage{nicefrac}       % compact symbols for 1/2, etc.
\usepackage{microtype}      % microtypography
\usepackage{xcolor}         % colors

\usepackage{xspace}
\usepackage{graphicx}
\usepackage{subcaption}
\usepackage{todonotes}
\usepackage{multirow}
\usepackage{makecell}
\usepackage{enumitem}
\usepackage{stfloats}
\fnbelowfloat

\def\jamba{Jamba\xspace}

\title{\jamba:\\ A Hybrid Transformer-Mamba Language Model
}

\author{%
Opher Lieber\thanks{Equal contribution.} \hspace{.75em} Barak Lenz\footnotemark[1] \hspace{.75em} Hofit Bata \hspace{.75em} Gal Cohen \hspace{.75em} Jhonathan Osin \hspace{.75em} \\ \bf Itay Dalmedigos  \hspace{.75em} \bf  Erez Safahi \hspace{.75em} Shaked Meirom \hspace{.75em} Yonatan Belinkov \\  \bf Shai Shalev-Shwartz \hspace{.75em} \bf Omri Abend \hspace{.75em}  \bf Raz Alon \hspace{.75em} Tomer Asida \\ \bf  Amir Bergman  \hspace{.75em} \bf Roman Glozman \hspace{.75em}  \bf  Michael Gokhman \hspace{.75em} Avshalom Manevich \\ \bf  Nir Ratner \hspace{.75em} \bf Noam Rozen \hspace{.75em} Erez Schwartz \hspace{.75em} Mor Zusman \hspace{.75em} Yoav Shoham \\
}

\begin{document}

\maketitle

\begin{center}
\vspace{-20pt}
\centering
\includegraphics[width=0.30\linewidth,keepaspectratio]{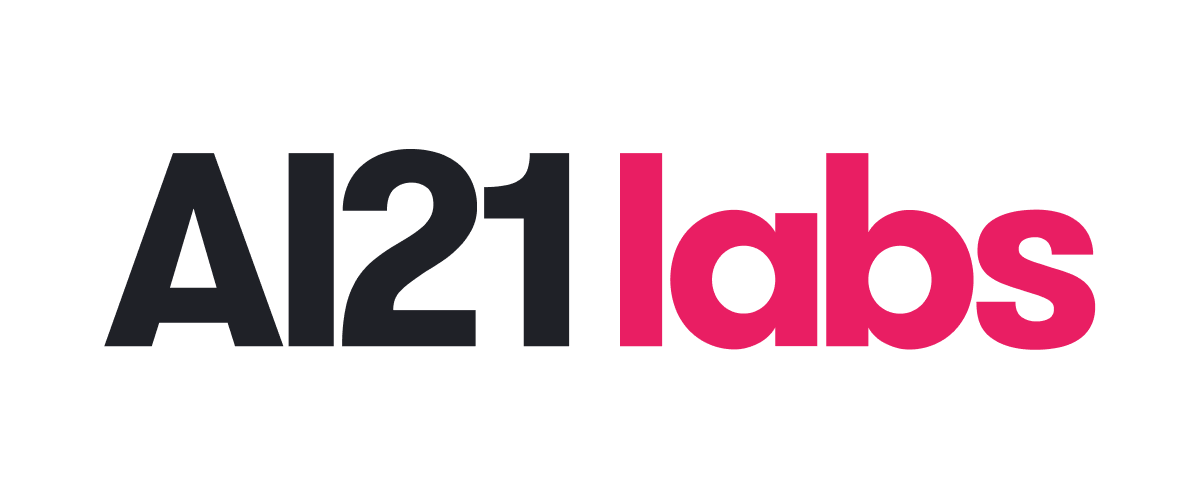}
\end{center}

\vspace{10pt}

\begin{abstract}
We present \jamba, a new base large language model based on a novel hybrid Transformer-Mamba mixture-of-experts (MoE) architecture. Specifically, \jamba interleaves blocks of Transformer and Mamba layers, enjoying the benefits of both model families. MoE is added in some of these layers to increase model capacity while keeping active parameter usage manageable. This flexible architecture allows resource- and objective-specific configurations. 
In the particular configuration we have implemented, we end up with a powerful model that fits in a single 80GB GPU. 
 Built at large scale, \jamba provides high throughput and small memory footprint compared to vanilla Transformers, and at the same time state-of-the-art performance on standard language model benchmarks and long-context evaluations. Remarkably, the model presents strong results for up to 256K tokens context length.
 We study various architectural decisions, such as how to combine Transformer and Mamba layers, and how to mix experts, and show that some of them are crucial in large scale modeling. We also describe several interesting properties of these architectures which the training and evaluation of \jamba have revealed, and plan to release  checkpoints from various ablation runs, to encourage further exploration of this novel architecture.
 We make the weights of our implementation of \jamba publicly available under a permissive license.
 \\\\
 \textbf{Model:} \url{https://huggingface.co/ai21labs/Jamba-v0.1}

\end{abstract}

\section{Introduction}
\label{sec:intro}

We introduce \jamba, a new publicly available large language model. \jamba is based on  a novel hybrid architecture, which combines Transformer layers \cite{vaswani2017attention} with Mamba layers \cite{gu2023mamba}, a recent state-space model \cite{gu2021combining,gu2021efficiently}, as well as a mixture-of-experts (MoE) module \cite{shazeer2016outrageously,fedus2022switch}. \jamba thus combines two orthogonal architectural designs that together give it improved performance and higher throughput, while maintaining a manageable memory footprint. The 7B-based \jamba model (12B active parameters, 52B total available parameters) we are releasing was designed to fit in a single 80GB GPU, but the \jamba architecture supports other design choices, depending on one's hardware and performance requirements.  

The fundamental novelty of \jamba is its hybrid Transformer-Mamba architecture (though see mention below of  recent related efforts). Despite the immense popularity of the Transformer as the predominant architecture for language models, it suffers from two main drawbacks. First, its high memory and compute requirements hinders the processing of long contexts, where the key-value (KV) cache size becomes a limiting factor. Second, its lack of a single summary state entails slow inference and low throughput, since each generated token performs a computation on the entire context. In contrast, older recurrent neural network (RNN) models, which summarize an arbitrarily long context in a single hidden state, do not suffer from these limitations. RNN models have their own shortcomings, however. They are costly to train since training cannot be parallelized across time steps. And they struggle with long distance relationships, which the hidden state captures to only a limited extent. 

Recent state space models (SSMs) like Mamba are more efficient to train than RNNs and are more capable at handling long distance relationships, but still lag behind the performance of comparably sized Transformer language models. Taking advantage of both model families, \jamba combines Transformer and Mamba layers, at a certain ratio. Varying the ratio of Transformer/Mamba layers allows balancing memory usage, efficient training, and long context capabilities.  

A few other recent attempts to combine Attention and SSM modules are worth noting. 
\cite{zuo2022efficient} mixes an S4 layer \cite{gu2021efficiently} with a local attention layer, followed by a sequence of local attention layers; it shows experiments with small models and simple tasks. 
\cite{gu2023mamba} reports that interleaving Mamba and attention layers is only slightly better than pure Mamba in terms of perplexity, with models up to 1.3B parameters. 
\cite{pilault2023block} starts with an SSM layer followed by chunk-based Transformers, with models up to 1.3B showing improved perplexity. 
\cite{fathullah23_interspeech} adds an SSM layer before the self-attention in a Transformer layer, while \cite{saon2023diagonal} adds the SSM after the self-attention, both showing improvements on speech recognition. 
\cite{park2024can} replaces the MLP layers in the Transformer by Mamba layers, and shows benefits in simple tasks. These efforts are different from \jamba both in the particular way in which the SSM component is mixed with the attention one, and in the scale of implementation. Closest are perhaps H3 \cite{fu2022hungry}, a specially designed SSM that enables induction capabilities, and a generalization called Hyena \cite{poli2023hyena}. The former proposed a hybrid architecture that replaces the second and middle layers with self-attention, and  was implemented with up to 2.7B parameters and 400B training tokens. However, as shown in \cite{gu2023mamba}, its perfomance lags that of pure Mamba.
Based on Hyena, StripedHyena \cite{stripedhyena} interleaves attention and SSM layers in a 7B parameter model. However, it lags behind the Attention-only Mistral-7B \cite{jiang2023mistral}. 
All of this renders \jamba the first production-grade Attention-SSM hybrid model.
Scaling the hybrid \jamba architecture required overcoming several obstacles, which we dicsuss in Section \ref{sec:ablations}. 

\jamba also includes MoE layers \cite{shazeer2016outrageously,fedus2022switch}, which allow increasing the model capacity (total number of available parameters) without increasing compute requirements (number of active parameters). MoE is a flexible approach that enables training extremely large models with strong performance \cite{jiang2024mixtral}. In \jamba, MoE is applied to some of the MLP layers. The more MoE layers, and the more experts in each MoE layer, the larger the total number of model parameters. In contrast, the more experts we use at each forward pass, the larger the number of active parameters as well as the compute requirement. In our implementation of \jamba, we apply MoE at  every other layer, with 16 experts and the top-2 experts used at each token (a more detailed discussion of the model architecture is provided below).

We evaluated our implementation of \jamba on a wide range of benchmarks and found it performs comparably to Mixtral-8x7B \cite{jiang2024mixtral}, which has a similar number of parameters, and also to the larger Llama-2 70B \cite{touvron2023llama}. In addition, our model supports a context length of 256K tokens -- the longest supported context length for production-grade publicly available models. On long-context evaluations,  \jamba outperformes Mixtral on most of the evaluated datasets. At the same time, \jamba is extremely efficient; for example, its  throughput is 3x that of Mixtral-8x7B for long contexts. Moreover, our model fits in a single GPU (with 8bit weights) even with contexts of over 128K tokens, which is impossible with similar-size attention-only models such as Mixtral-8x7B.    

Somewhat unusual for a new architecture, we release \jamba (12B active parameters, 52B total available parameters) under Apache 2.0 license: \url{https://huggingface.co/ai21labs/Jamba-v0.1}. We do so since we feel that the novel architecture of \jamba calls for further study, experimentation, and optimization by the community. Our design was based on various ablation experiments we conducted to explore the effect of different tradeoffs and design choices, and insights gleaned from those. These ablations were performed at scales of up to 7B parameters, and training runs of up to 250B tokens. We plan to release model checkpoints from these runs. 

\emph{\textbf{Important notice}: The \jamba model released is a pretrained \textbf{base} model, which did not go through alignment or instruction tuning, and does not have moderation mechanisms. It should not be used in production environments or with end users without additional adaptation.}

\section{Model Architecture} \label{sec:arch}

\begin{figure}
    \centering
    \includegraphics[width=.9\textwidth]{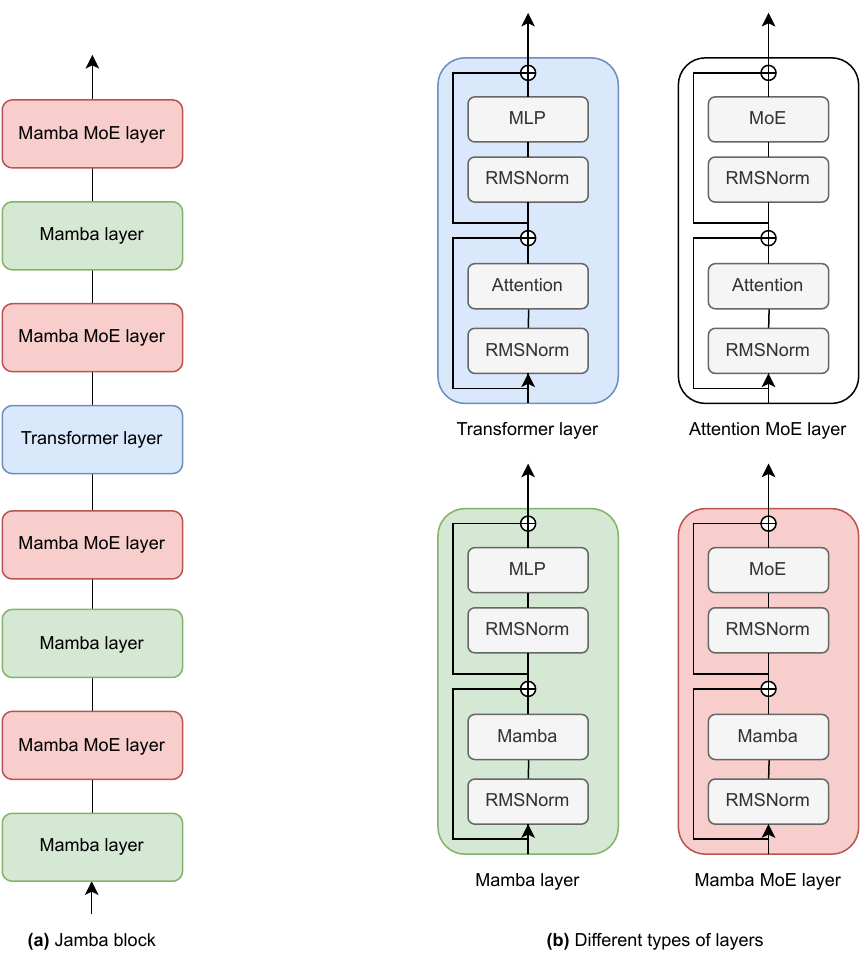}
    \caption{\textbf{(a)} A single Jamba block. \textbf{(b)} Different types of layers. The implementation shown here is with $l = 8$, $a:m = 1:7$ ratio of attention-to-Mamba layers, and MoE applied every $e = 2$ layers.}
    \label{fig:jamba-arch}
\end{figure}

\jamba is a hybrid decoder architecture that mixes Transformer layers \cite{vaswani2017attention} with Mamba layers \cite{gu2023mamba}, a recent state-space model (SSM) \cite{gu2021combining,gu2021efficiently}, in addition to a mixture-of-experts (MoE) module \cite{shazeer2016outrageously,fedus2022switch}. We call the combination of these three elements a Jamba block. See Figure \ref{fig:jamba-arch} for an illustration. 

Combining Transformer, Mamba, and MoE elements allows flexibility in balancing among the sometimes conflicting objectives of low memory usage, high throughput, and high quality. In terms of memory usage, note that comparing the total size of the model parameters can be misleading. In an MoE model, the number of active parameters that participate in any given forward step may be much smaller than the total number of parameters. Another important consideration is the KV cache -- the memory required to store the attention keys and values in the context. When scaling Transformer models to long contexts, the KV cache becomes a limiting factor. Trading off attention layers for Mamba layers reduces the total size of the KV cache. Our architecture aims to provide not only a small number of active parameters but also an 8x smaller KV cache compared to a vanilla Transformer. 
Table \ref{table:params-cache} compares Jamba with recent publicly available models, showing its advantage in maintaining a small KV cache even with 256K token contexts.

\begin{table}[h]
\centering
\begin{tabular}{l c c c}
    \toprule
    & Available params & Active params & KV cache (256K context, 16bit) \\ 
    \midrule     
    LLAMA-2 & 6.7B & 6.7B & 128GB \\
    Mistral & 7.2B & 7.2B & 32GB \\ 
    Mixtral & 46.7B & 12.9B & 32GB \\ 
    Jamba & 52B & 12B & 4GB \\
    \bottomrule
\end{tabular}
\vspace{1pt}
\caption{Comparison of Jamba and recent open models in terms of total available parameters, active parameters, and KV cache memory on long contexts. Jamba provides a substantial reduction in the KV cache memory requirements.}
\label{table:params-cache}
\end{table}

In terms of throughput, with short sequences, attention operations take up a small fraction of the inference and training FLOPS \cite{chowdhery2023palm}. However, with long sequences, attention hogs most of the compute. In contrast, Mamba layers are more compute-efficient. Thus, increasing the ratio of Mamba layers improves throughput especially for long sequences.  

Here is a description of the main configuration, which provides improved performance and efficiency. Section \ref{sec:ablations} contains results from ablation experiments supporting the design choices. 

The basic component is a Jamba block, which may be repeated in sequence. Each Jamba block is a combination of Mamba or Attention layers. Each such layer contains either an attention or a Mamba module, followed by a multi-layer perceptron (MLP). The different possible types of layers are shown in Figure \ref{fig:jamba-arch}(b).\footnote{The figure shows a potential Attention MoE layer, which our architecture does not use, but future variants could.} A Jamba block contains $l$ layers, which are mixed at a ratio of $a:m$, meaning $a$ attention layers for every $m$ Mamba layers. 

In Jamba, some of the MLPs may be replaced by MoE layers, which helps increase the model capacity while keeping the active number of parameters, and thus the compute, small. The MoE module may be applied to MLPs every $e$ layers. When using MoE, there are $n$ possible experts per layer, with a router choosing the top $K$ experts at each token.
In summary, the different degrees of freedom in the Jamba architecture are:
\begin{itemize}
\item $l$: The number of layers.
\item $a:m$: ratio of attention-to-Mamba layers.
\item $e$: how often to use MoE instead of a single MLP.     
\item $n$: total number of experts per layer. 
\item $K$: number of top experts used at each token. 
\end{itemize}

Given this design space, Jamba provides  flexibility in preferring certain properties over others. For example, increasing $m$ and decreasing $a$, that is, increasing the ratio of Mamba layers at the expense of attention layers, reduces the required memory for storing the key-value cache. This reduces the overall memory footprint, which is especially important for processing long sequences. Increasing the ratio of Mamba layers also improves throughput, especially at long sequences. However, decreasing $a$ might lower the model’s capabilities. 

Additionally, balancing $n$, $K$, and $e$ affects the relationship between active parameters and total available parameters. A larger $n$ increases the model capacity at the expense of memory footprint, while a larger $K$ increases the active parameter usage and the compute requirement. In contrast, a larger $e$ decreases the model capacity, while decreasing both compute (when $K$>1) and memory requirements, and allowing for less communication dependencies (decreasing memory transfers as well as inter-GPU communication during expert-parallel training and inference).

\jamba's implementation of Mamba layers incorporate several normalizations that help stabilize training in large model scales. In particular, we apply RMSNorm \cite{zhang2019root} in the Mamba layers. 

We found that with the Mamba layer, positional embeddings or mechanisms like RoPE \cite{su2024roformer} are not necessary, and so we do not use any explicit positional information. 

Other architecture details are standard, including grouped-query attention (GQA), SwiGLU activation function \cite{shazeer2020glu,chowdhery2023palm,touvron2023llama}, and load balancing for the MoE \cite{fedus2022switch}. The vocabulary size is 64K. The tokenizer is trained with BPE \cite{gage1994new,sennrich2016neural,mielke2021between} and each digit is a separate token \cite{chowdhery2023palm}.  We also remove the dummy space used in Llama and Mistral tokenizers for more consistent and reversible tokenization.

\section{Reaping the Benefits}

\subsection{Jamba Implementation for a Single 80GB GPU} \label{sec:impl-single-gpu}

The specific configuration in our implementation was chosen to fit in a single 80GB GPU, while achieving best performance in the sense of quality and throughput. In our implementation we have a sequence of 4 Jamba blocks. Each Jamba block has the following configuration:
\begin{itemize}
\item $l = 8$: The number of layers.
\item $a:m = 1:7$: ratio attention-to-Mamba layers.
\item $e = 2$: how often to use MoE instead of a single MLP. 
\item $n = 16$: total number of experts. 
\item $K = 2$: number of top experts used at each token. 
\end{itemize}

The $a:m = 1:7$ ratio was chosen according to preliminary ablations, as shown in Section \ref{sec:ablations}, since this ratio was the most compute-efficient variant amongst the best performing variants in terms of quality.

The configuration of the experts was chosen to enable the model to fit in a single 80GB GPU (with int8 weights), while including sufficient memory for the inputs. In particular, $n$ and $e$ were balanced to have an average of $\sim$8 experts per layer. In addition, we balanced $n$, $K$, and $e$ to allow for high quality, while keeping both compute requirements and communication dependencies (memory transfers) checked. Accordingly, we chose to replace the MLP module with MoE on every other layer, as well as have a total of 16 experts, two of which are used at each token. These choices were inspired by prior work on MoE \cite{zoph2022st,clark2022unified} and verified in preliminary experiments.

\begin{figure}[h]
    \centering
    \includegraphics[width=.6\textwidth]{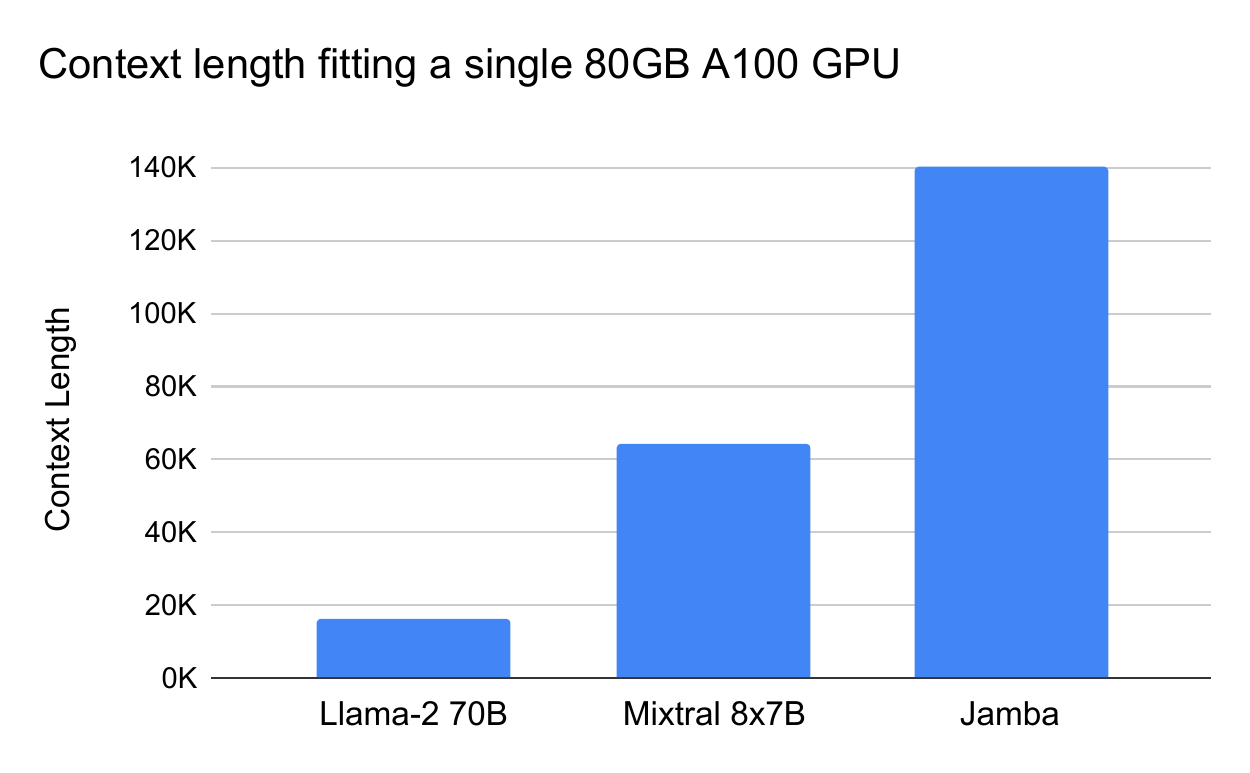}
    \caption{Comparison of maximum context length fitting in a single A100 80GB GPU. Jamba enables 2x the context length of Mixtral and 7x that of Llama-2-70B.}
    \label{fig:single-gpu-context}
\end{figure}

Figure \ref{fig:single-gpu-context} shows the maximal context length that fits a single 80GB GPU with our Jamba implementation compared to Mixtral 8x7B and Llama-2-70B. Jamba provides 2x the context length of Mixtral and 7x that of Llama-2-70B.

Overall, our Jamba implementation was successfully trained on context lengths of up to 1M tokens. The released model supports lengths of up to 256K tokens. 

\subsection{Throughput Analysis} \label{sec:throughput}
\vspace{-5pt}

\begin{figure}[h]
    \centering
    \begin{subfigure}{0.49\textwidth}
    \includegraphics[width=\linewidth]{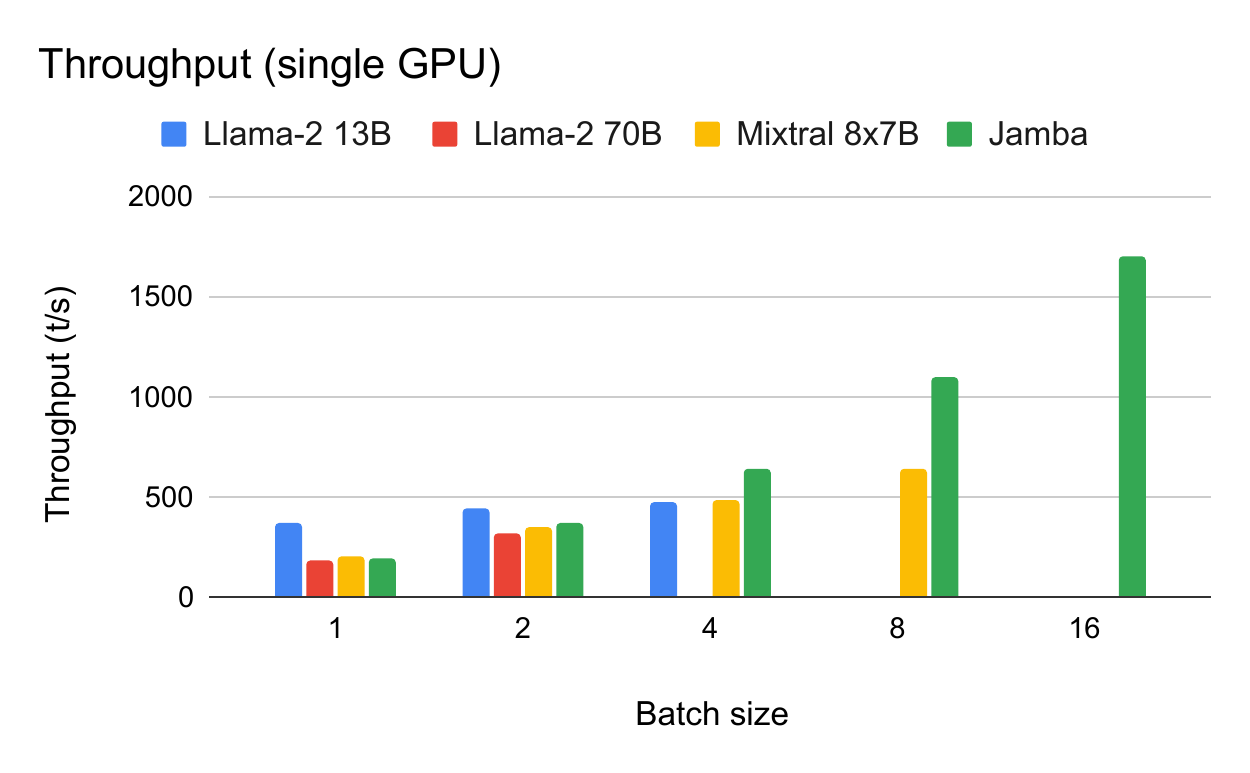}
    \caption{Throughput at different batch sizes (single A100 GPU, 8K context length). Jamba allows processing large batches, with a throughput 3x greater than Mixtral.}
    \label{fig:throughput-1}
    \end{subfigure}\hfill%
    \begin{subfigure}{0.49\textwidth}
    \includegraphics[width=\linewidth]{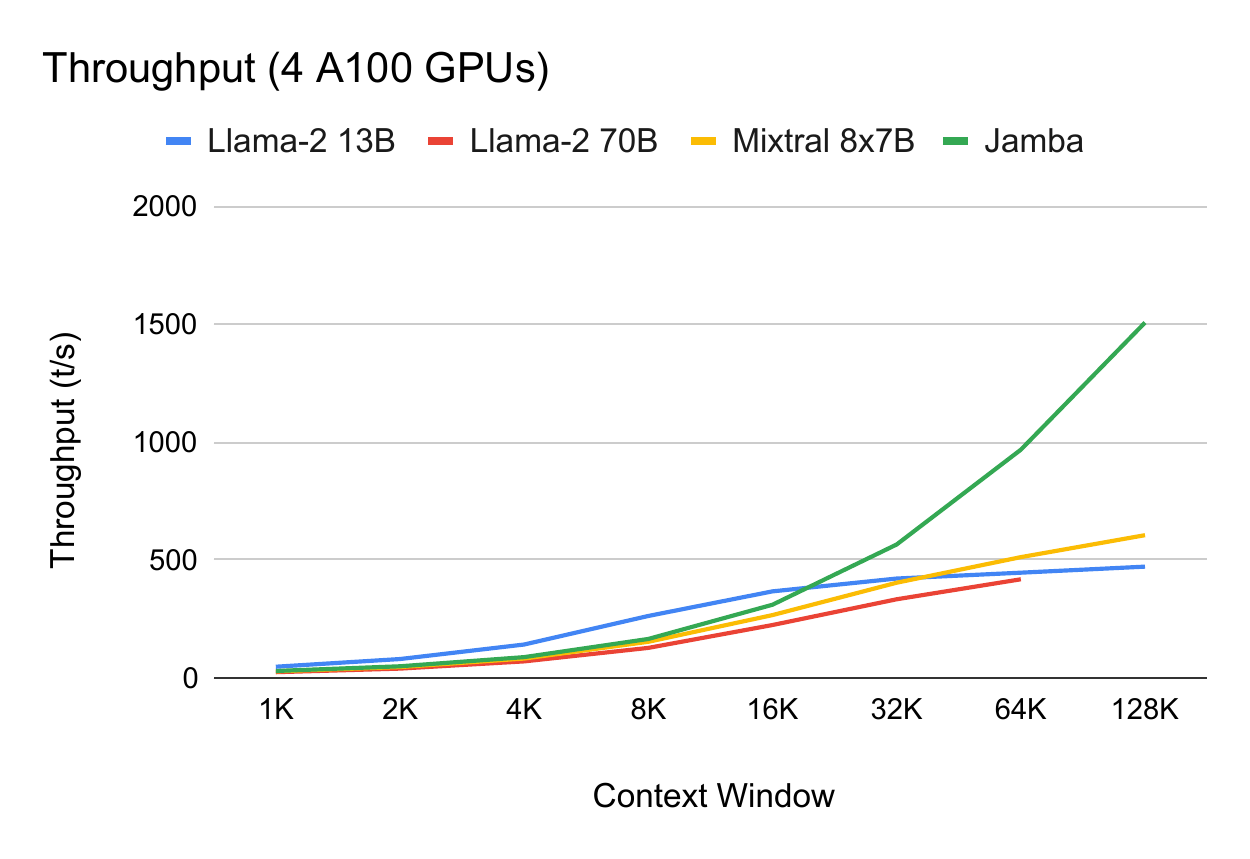}
    \caption{Throughput at different context lengths (single batch, 4 A100 GPUs).  With a context of 128K tokens, Jamba obtains 3x the throughput of Mixtral, while Llama-2-70B does not fit with this long context.}
    \label{fig:throughput-4}
    \end{subfigure}    
    \caption{Comparison of throughput (tokens/second) with Jamba and recent open models.}
    \label{fig:enter-label}
\end{figure}

For concreteness, we present results of the throughput in two specific settings.\footnote{Referring to end-to-end throughput (encoding+decoding). The results should be taken relatively rather than absolutely, as they are without possible optimizations.} 
In the first setting, we have varying batch size, a single A100 80 GB GPU, int8 quantization, 8K context length, generating output of 512 tokens. 
As Figure \ref{fig:throughput-1}  shows, Jamba allows processing of large batches, leading to a 3x increase in throughput (tokens/second) over Mixtral, which does not fit with a batch of 16 despite having a similar number of active parameters. 

In the second setting, we have a single batch, 4 A100 GPUs, no quantization, varying context lengths, generating output of 512 tokens. 
As demonstrated in Figure \ref{fig:throughput-4}, at small context lengths all models have a similar throughput. Jamba excels at long contexts; with 128K tokens its throughput is 3x that of Mixtral. Note that this is despite the fact that \jamba has not yet enjoyed  optimizations of the kind the community has developed for  pure Transformer models over the past six years. We can expect the throughut gap to increase as such optimizations are developed also for \jamba.

\vspace{-10pt}
\section{Training Infrastructure and Dataset}
\vspace{-5pt}

The model was trained on NVIDIA H100 GPUs. We used an in-house proprietary framework allowing efficient large-scale training including FSDP, tensor parallelism, sequence parallelism, and expert parallelism. 

\jamba is trained on an in-house dataset that contains text data from the Web, books, and code, with the last update in March 2024. Our data processing pipeline includes quality filters and deduplication. 

\vspace{-5pt}
\section{Evaluation} \label{sec:evaluation}
\vspace{-5pt}

In general we approach benchmarks cautiously, as they correlate only partially with what matters in real applications, and furthermore invite gaming the system in order to boast vanity numbers. Nevertheless, we present several indicative results.

\vspace{-5pt}
\subsection{Academic Benchmarks} \label{sec:evaluation-benchmarks}

We report results with a wide range of  standard academic benchmarks:  

\begin{description}[itemsep=1pt,parsep=1pt,topsep=1pt] 
    \item[Common sense reasoning:] HellaSwag (10-shot) \cite{zellers2019hellaswag}, WinoGrande (5-shot) \cite{sakaguchi2020winogrande}, ARC-E (0-shot) and ARC-Challenge (25-shot) \cite{clark2018think}, and PIQA (zero-shot) \cite{bisk2020piqa}.
    \item[Reading Comprehension:] BoolQ (10-shots) \cite{clark2019boolq} and QuAC (zero-shot) \cite{choi2018quac}.
    \item[Others:] GSM8K (3-shot CoT) \cite{cobbe2021training}, HumanEval (pass@1)  \cite{chen2021evaluating}, Natural Questions closed-book (NQ; 5-shot) \cite{kwiatkowski2019natural}, and  TruthfulQA (zero-shot) \cite{lin-etal-2022-truthfulqa}. 
    \item[Aggregate benchmarks:] MMLU (5-shot) \cite{hendrycks2020measuring} and BBH (3-shot) \cite{suzgun2023challenging}. 
\end{description}

Table \ref{table:eval-benchmarks} compares \jamba to several publicly available models on common academic benchmarks for evaluating language models. We compare with Llama-2 13B \cite{touvron2023llama}, which has about the same number of active paramters as our model, Llama-2 70B, which is larger than our model, Gemma \cite{team2024gemma}, which has 7B parameters, and Mixtral \cite{jiang2024mixtral}, which has about the same number of active and total parameters as our model. 

In most tasks, \jamba performs comparably to leading publicly available models of similar or larger size, including Llama-2 70B and Mixtral. At the same time, our model has a smaller number of total available parameters than Llama-2 (52B compared to 70B). Moreover, as a sparse model, \jamba has only 12B active parameters, similar to Mixtral's 12.9B active parameters. However, as a fully-attentional model, Mixtral has a large memory footprint with long sequences, requiring 32GB for the KV cache with 256K tokens. In contrast, thanks to its hybrid Attention-Mamba architecture, \jamba's KV cache takes only 4GB even at such a long context (Section \ref{sec:arch}).
Importantly, our \jamba achieves such a strong performance while having much better throughput than Llama-2 70B and Mixtral, up to 3x improvement (Section \ref{sec:throughput}). 

In summary, \jamba demostrates the ability of hybrid architectures to reach the performance of state-of-the-art Transformer models of the same size class, while having the benefits of an SSM.

\begin{table}[t]
    \centering
    \resizebox{\textwidth}{!}{%
    \begin{tabular}{l ccccccc}
    \toprule
  & \multicolumn{5}{c}{\bf Reasoning}  \\ \cmidrule(lr){2-6} 
  &  \multicolumn{1}{c}{\bf HellaSwag}  &  \multicolumn{1}{c}{\bf WinoGrande}  &  \multicolumn{1}{c}{\bf ARC-E}  &  \multicolumn{1}{c}{\bf ARC-C}  &  \multicolumn{1}{c}{\bf PIQA}  &  \multicolumn{1}{c}{\bf NQ}  &  \multicolumn{1}{c}{\bf TruthfulQA} \\ 
\midrule 
\textbf{Llama-2 13B}  & 80.7 & 72.8 & 77.3 & 59.4 & 80.5 &  37.7   &  37.4 \\ 
\textbf{Llama-2 70B}  & 85.3 & 80.2 & 80.2 &  \bf 67.3  & 82.8 &  \bf 46.9   &  44.9 \\ 
\textbf{Gemma}  & 81.2 & 72.3 & \bf 81.5 & 53.2 & 81.2 &  32.6  &  44.8 \\ 
\textbf{Mixtral}  & 86.7 & 81.2 & 77.6 & 66 & 83 &  44.8   &  \bf 46.8 \\ 
\midrule 
\textbf{Jamba}  &  \bf 87.1  &  \bf 82.5  & 73.5 & 64.4 &  \bf 83.2  &  45.9   &  46.4 \\ 
\midrule 
\\ 
& \multicolumn{2}{c}{\bf Comprehension} &  &  & \multicolumn{2}{c}{\bf Aggregate} \\
\cmidrule(lr){2-3}  \cmidrule(lr){6-7} 
 &  \multicolumn{1}{c}{\bf BoolQ}  &  \multicolumn{1}{c}{\bf QuAC}  &  \multicolumn{1}{c}{\bf GSM8K}  &  \multicolumn{1}{c}{\bf HumanEval}  &  \multicolumn{1}{c}{\bf MMLU}  &  \multicolumn{1}{c}{\bf BBH}  \\ 
\cmidrule(lr){1-7} 
\textbf{Llama-2 13B}  & 81.7 &  \bf 42.7  & 34.7 & 18.3 & 54.8 & 39.4 \\ 
\textbf{Llama-2 70B}  & 85 & 42.4 & 55.3 & 29.9 & 69.8 & 51.2 \\ 
\textbf{Gemma}  & 87.2 & 39.2 & 54.5 & 32.3 & 64.3 &  \bf 55.1  \\ 
\textbf{Mixtral}  &  \bf 88.4  & 40.9 &  \bf 60.4  &  \bf 34.8  &  \bf 70.6  & 50.3 \\ 
\cmidrule(lr){1-7} 
\textbf{Jamba}  & 88.2 & 40.9 & 59.9 & 29.3 & 67.4 & 45.4 \\ 
    \bottomrule      
    \end{tabular}}
    \vspace{1pt}
    \caption{Comparison of Jamba with other publicly available models. Jamba obtains similar performance with much better throughput. }
    \label{table:eval-benchmarks}
    \vspace{-5pt}
\end{table}

\subsection{Long-Context Evaluations} \label{sec:evaluation-long}

We have successfully trained Jamba models with context lengths of up to 1M tokens. The released model handles context lengths of up to 256K tokens. In this section, we evaluate it on synthetic and naturalistic benchmarks that test its long-context capabilities. 

\subsubsection{Needle-in-a-haystack} \label{sec:evaluation-long-needle}

\begin{figure}
    \centering
    \includegraphics[width=.91\textwidth]{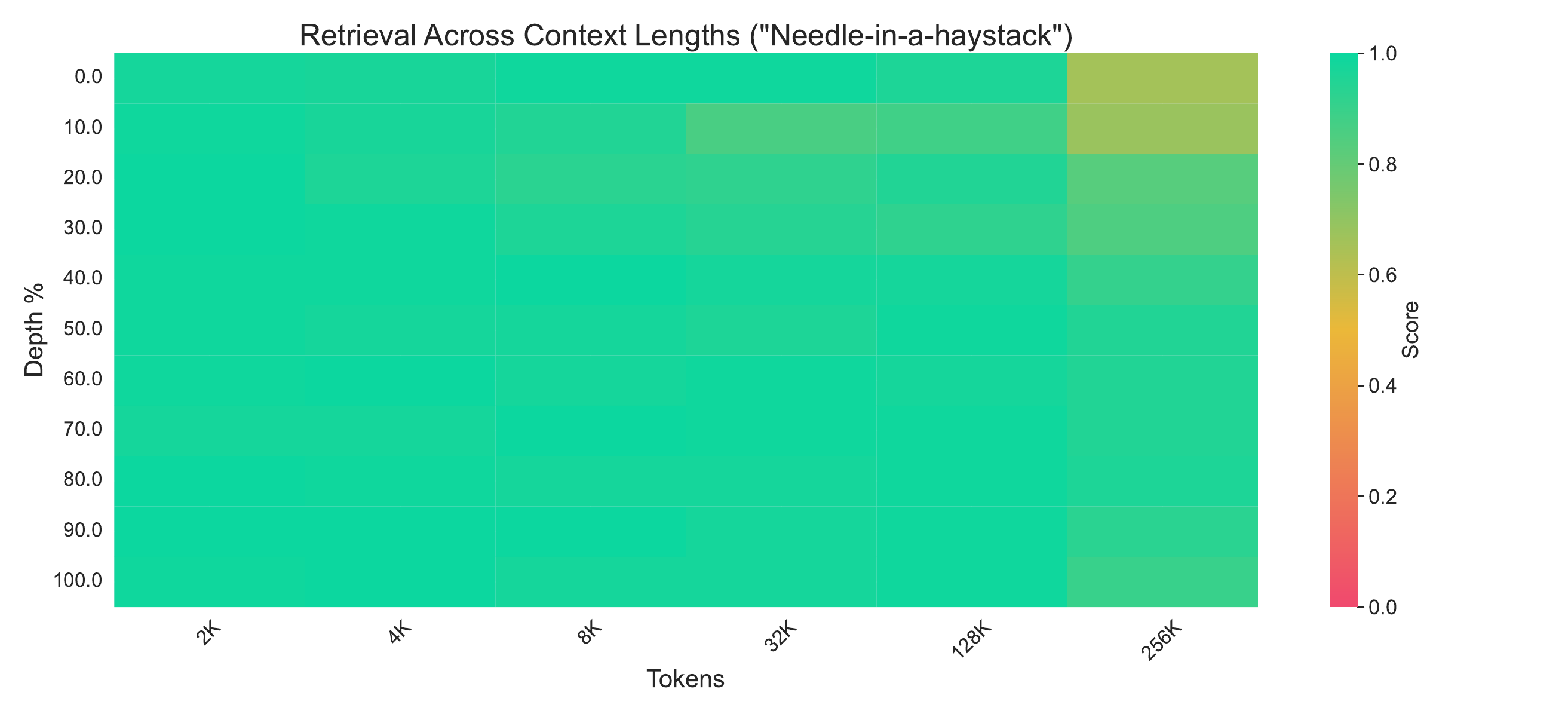}
    \caption{A needle-in-a-haystack evaluation showing Jamba's ability to recall statements placed in the middle of contexts of up to 256K tokens length.}
    \label{fig:needle}
    \vspace{-3pt}
\end{figure}

As Figure \ref{fig:needle} shows, Jamba has excellent performance in the needle-in-a-haystack evaluation, which requires retrieving a simple statement planted in a long context window \cite{kamradt2023}. This result is noteworthy especially given that our implementation of Jamba uses only 4 attention layers. 

\subsubsection{Naturalistic long-context evaluation} \label{sec:evaluation-long-naturalistic}

\begin{figure}[h]
    \centering
    \begin{subfigure}{0.49\textwidth}
    \includegraphics[width=\linewidth]{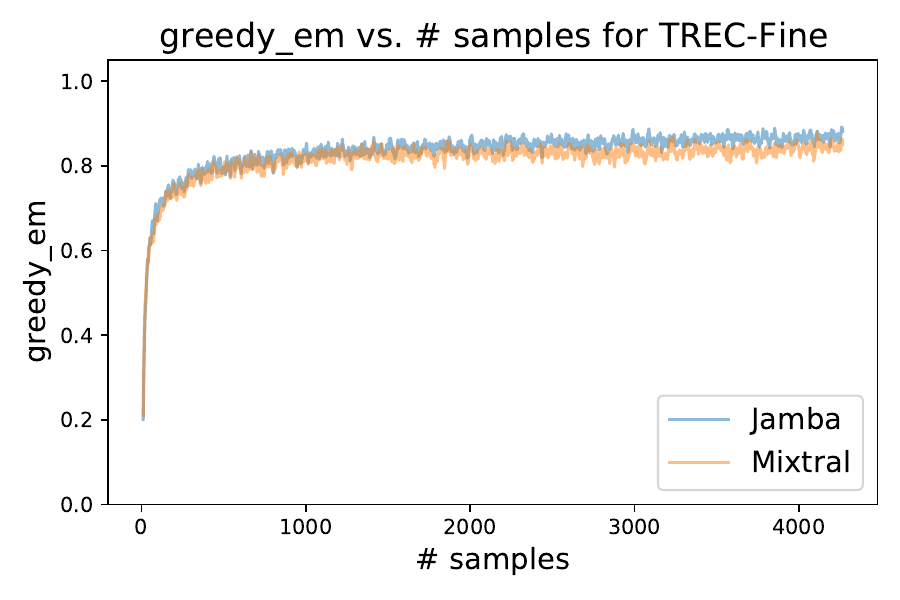}
    \caption{TREC Fine-grained.}
    \label{fig:long-classification-trec}
    \end{subfigure}\hfill%
    \begin{subfigure}{0.49\textwidth}
    \includegraphics[width=\linewidth]{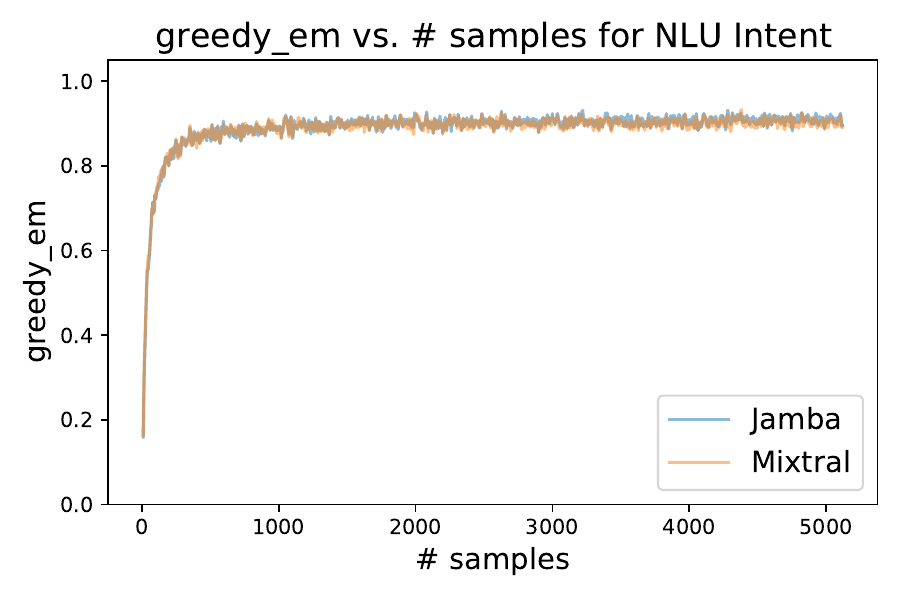}
    \caption{NLU Intent.}
    \label{fig:long-classification-nlu}
    \end{subfigure}\\
    \begin{subfigure}{0.49\textwidth}
    \includegraphics[width=\linewidth]{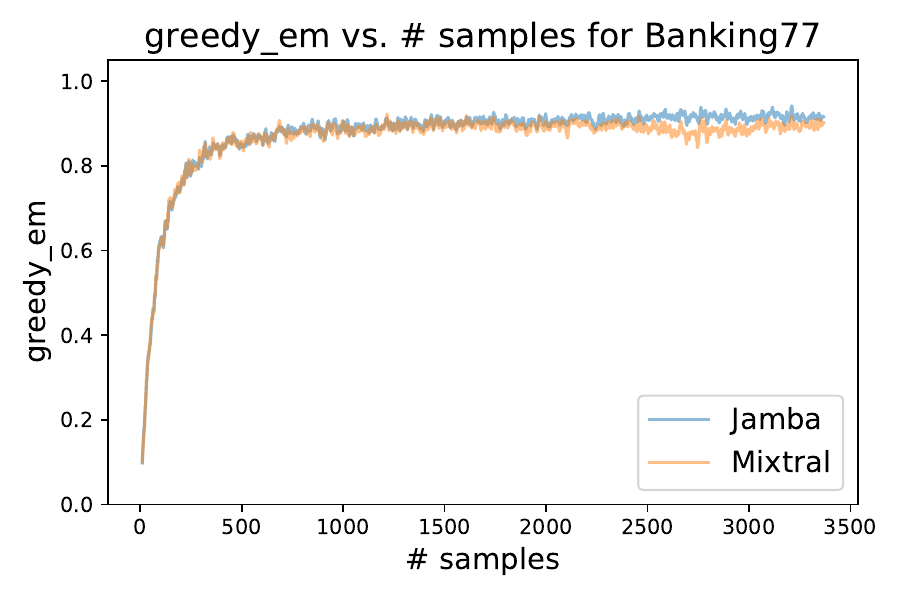}
    \caption{Banking77.}
    \label{fig:long-classification-banking}
    \end{subfigure}\hfill 
    \begin{subfigure}{0.49\textwidth}
    \includegraphics[width=\linewidth]{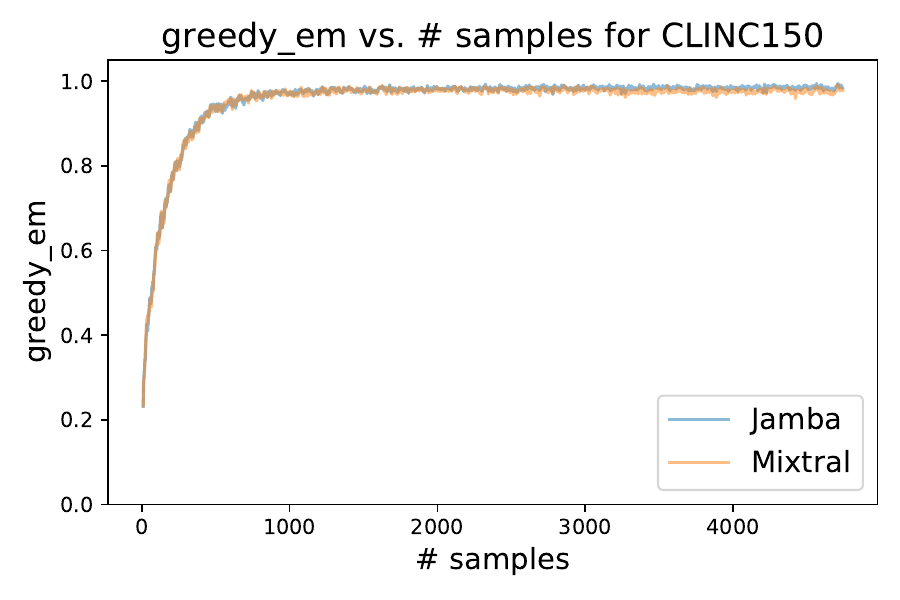}
    \caption{CLINC150.}
    \label{fig:long-classification-clinc}
    \end{subfigure}        
    \caption{Comparison of Jamba and Mixtral on few-shot classification with a large number of few-shot examples. Results are exact match with greedy decoding.}
    \label{fig:long-context-classification}
\end{figure}

We evaluate Jamba’s ability to handle long contexts in two settings.
First, we evaluate the model on several classification tasks commonly used for assessing in-context learning, with an increasing number of few-shot examples. In particular, we use the four datasets with the largest label space from \cite{ratner2023parallel}, which showed that such tasks benefit most from using more few-shot examples: Trec-Fine (fine-grained question type classification, 50 labels; \cite{li2002learning}), NLU Intent (intent classification in natural language understanding, 68 labels; \cite{liu2021benchmarking}), Banking77 (intent classification in the banking domain, 77 labels; \cite{casanueva2020efficient}),  and CLINC150 (intent classification, 150 labels; \cite{larson2019evaluation}). In each case we add examples up to a context length of 128K tokens. 
Figure \ref{fig:long-context-classification} shows the results, comparing \jamba to Mixtral. In Trec-Fine and Banking77, \jamba outperforms Mixtral, especially with a large number of few-shot examples. In NLU Intent and CLINC150, the two models are on par. 

Second, we use question-answering datasets, consisting of long inputs. To this end, we repurpose five of the longest-context datasets from L-Eval \cite{an2023eval}, by structuring them in a few-shot format (we use 3-shots in these experiments). Specifically, we evaluated the models on the following datasets: NarrativeQA (QA on narratives; \cite{kocisky2018narrativeqa}), LongFQA (finance; \cite{an2023eval}), Natural Questions (NQ; Wikipedia; \cite{kwiatkowski2019natural}), CUAD (law; \cite{hendrycks2021cuad}), and SFiction (science fiction). The average input length in these datasets ranges from 6K to 62K tokens. These lengths are further highly expanded by the few-shot format. 

Table \ref{table:leval} summarizes the results, in terms of F1.\footnote{F1 score is the recommended metric in L-Eval \cite{an2023eval}. In addition, our setup calibrates the length of the few-shot completions to approximately match the length of the test completion,  increasing the credibility of F1.} \jamba outperforms Mixtral on most of the datasets as well as on average. In addition, as these long-context tasks require substantial computation, here \jamba's efficiency shines, with much better throughput with long contexts (Section \ref{sec:throughput}). 

\begin{table}[h]
    \centering
    \begin{tabular}{l ccccc | c}
    \toprule 
    & LongFQA & CUAD & NarrativeQA & NQ & SFiction &  Avg \\ 
    \midrule 
    Mixtral & 0.42 & 0.46 & 0.29 &  0.58 & 0.42 & 0.43 \\ 
    Jamba & 0.44 & 0.44 & 0.30 & 0.60 & 0.40 & 0.44 \\ 
    \bottomrule 
    \end{tabular}
    \vspace{1pt}
    \caption{Results (F1) on long-context QA benchmarks, with a 3-shot format.}
    \label{table:leval}
\end{table}

\vspace{-20pt}
\section{Ablations and Insights} \label{sec:ablations}

This section discusses ablation experiments we ran for different design choices in our implementation of the Jamba architecture. First we show the benefit of combining attention and Mamba layers, at which ratio they should be combined, and how to interleave them. 
We investigate cases where pure Mamba fails, suggesting that it struggles to develop in-context learning capabilities, while the Attention-Mamba hybrid exhibits in-context learning similar to vanilla Transformers. 
Then we show the benefit of adding MoE on top of a hybrid Attention-Mamba model. Finally, we share two additional learnings that we found useful: explicit positional information is not needed in \jamba, and Mamba layers necessitate special normalization to stabilize training at large scale.\footnote{In all the ablations, ``pure Mamba'' refers to models with Mamba layers interleaved with MLP layers.}

For these ablations, we report the following measures, which exhibit informative performance even at small data or model scale. 
\begin{itemize}[itemsep=2pt,topsep=2pt,parsep=2pt]
    \item Academic benchmarks: HellaSwag (10-shot) \cite{zellers2019hellaswag}, WinoGrande (5-shot) \cite{sakaguchi2020winogrande}, Natural Questions closed-book (NQ; 5-shot) \cite{kwiatkowski2019natural}.
    \item HuggingFace OpenLLM leaderboard (OLLM) \cite{ollm}: a summary statistic of several datasets. We report results with our reproduction.
    \item Perplexity evaluations: we report log-prob (per byte) on texts from three domains: C4, Books, and code.  
\end{itemize}

\subsection{Benefits of combining Attention and Mamba} 

\begin{table}[b]
    \centering
    \begin{tabular}{l c c c c c c c}
    \toprule 
         & & \multirow{2}{*}[-5pt]{\makecell[b]{Hella\\Swag}} & \multirow{2}{*}[-5pt]{\makecell[b]{Wino\\Grande}}  & & \multicolumn{3}{c}{log-prob} \\ 
         \cmidrule(lr){6-8}
         & OLLM &  &  & NQ &  C4 &  Books &  Code \\
         \midrule 
        Attention & 36.4  & 62.4 & 59.6 & 14.5 & -0.543 & -0.659 & -0.331 \\ 
        Mamba & 36.1 & 62.6 & 59.4 & 14.5 & -0.543 & -0.661 & -0.334 \\ 
        Jamba ($a:m = 1:3$, no MoE) &  37.2 & 65.1 & 61.7 & 16.5 & -0.533 & -0.649 & -0.321 \\       
        Jamba ($a:m = 1:7$, no MoE) &  37.2 & 65.1 & 61.7 & 16.0 & -0.533 & -0.650 & -0.321 \\ 
        \bottomrule 
    \end{tabular}
    \hspace{1pt}
    \caption{Results on academic benchmarks and log-probability evaluations showing an improved performance of Attention-Mamba (no MoE) compared to vanilla Attention and Mamba models. There is no substantial difference between 1:3 and 1:7  ratios of Attention-to-Mamba layers.  Models are 1.3B parameters, trained for 250B tokens. }
    \label{table:attn-mamba-amba-1.3b}
\end{table}

\begin{figure}[h]
    \centering
    \includegraphics[width=.75\textwidth]{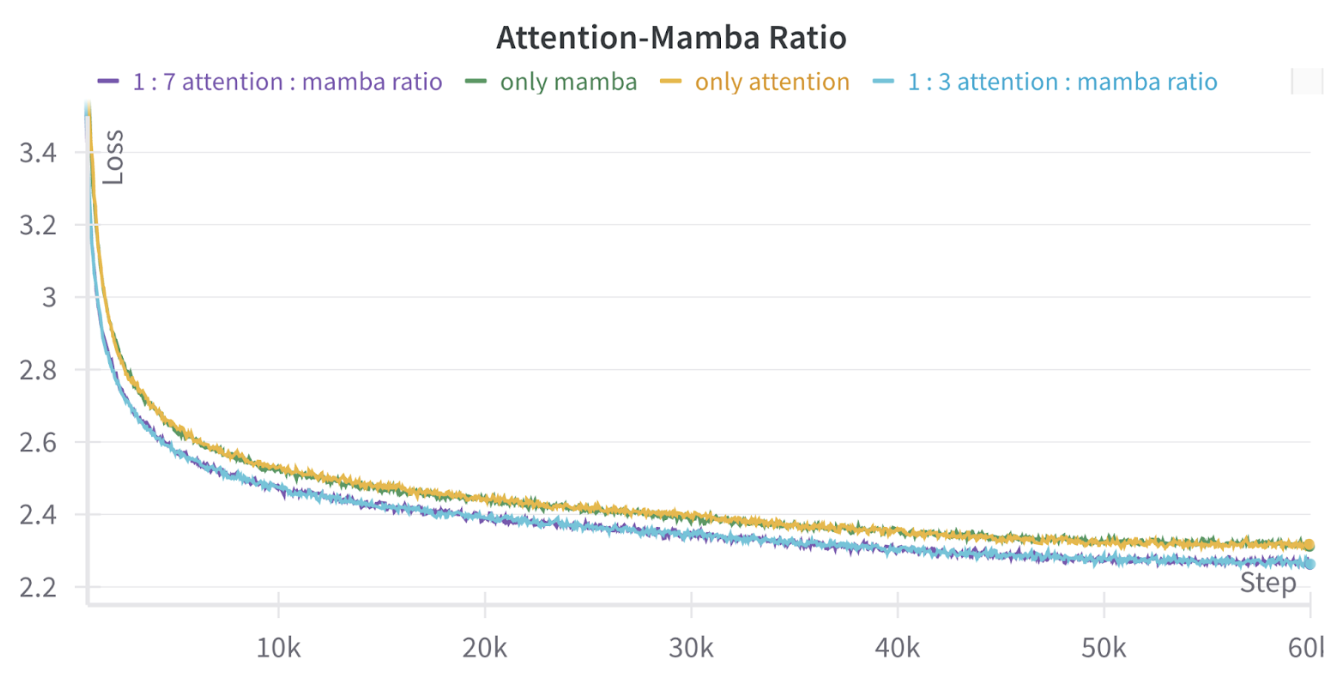}
    \caption{Training loss curves for pure Attention, pure Mamba, and Attention-Mamba hybrids (no MoE), with ratios $a:m$ of 1:3 and 1:7. All models are 1.3B parameters. The two hybrids achieve better loss throughout this training run, without any noticeable difference between the different Attention/Mamba ratios. }
    \label{fig:loss-attn-mamba-jamba-ratio}
\end{figure}

We first investigate the ratio of Attention to Mamba layers ($a:m$), with 1.3B parameters models trained for 250B tokens. As Table \ref{table:attn-mamba-amba-1.3b} shows, the hybrid Jamba model outperforms the pure attention or Mamba models. The ratio of attention-to-Mamba layers may be 1:3 or 1:7 with virtually no performance difference. 
Figure \ref{fig:loss-attn-mamba-jamba-ratio} shows the training loss of these models, where Jamba exhibits improved loss during training.
Given that a 1:7 ratio is more compute-efficient and shows similar performance, we opt for it in our larger-scale experiments.

Next, we compare  performance of vanilla Transformer, vanilla Mamba, and Attention-Mamba hybrid models, at 7B model size, after training on 50B tokens. As Table \ref{table:attn-mamba-amba-7b} shows, the pure Mamba model is quite competitive, but lags slightly behind pure Attention. The hybrid Attention-Mamba (without MoE) outperforms the pure models while obtaining better throughput than the vanilla Transformer (Section \ref{sec:throughput}).

\begin{table}[h]
    \centering
    \begin{tabular}{l c c c c c c c}
    \toprule 
         & & \multirow{2}{*}[-5pt]{\makecell[b]{Hella\\Swag}} & \multirow{2}{*}[-5pt]{\makecell[b]{Wino\\Grande}}  & & \multicolumn{3}{c}{log-prob} \\ 
         \cmidrule(lr){6-8}
         & OLLM &  &  & NQ &  C4 &  Books &  Code \\
         \midrule 
        Attention & 36.1 & 60.4 & 59.7 & 13.7 & -0.555 & -0.666 & -0.347 \\ 
        Mamba & 35.3 &  60.2 & 55.8 & 14.0 & -0.554 & -0.667 & -0.355 \\ 
        Jamba ($a:m = 1:7$, no MoE) &  36.6 &  62.5 & 58.8 & 15.4 & -0.547 & -0.658 & -0.340 \\ 
        \bottomrule 
    \end{tabular}
    \vspace{1pt}
    \caption{Results on academic benchmarks and log-prob evaluations, comparing pure Attention, pure Mamba, and Attention-Mamba hybrid (no MoE). Models are 7B parameters, trained for 50B tokens. }
    \label{table:attn-mamba-amba-7b}
\end{table}

\begin{figure}[h]
    \centering
    \includegraphics[width=.75\textwidth]{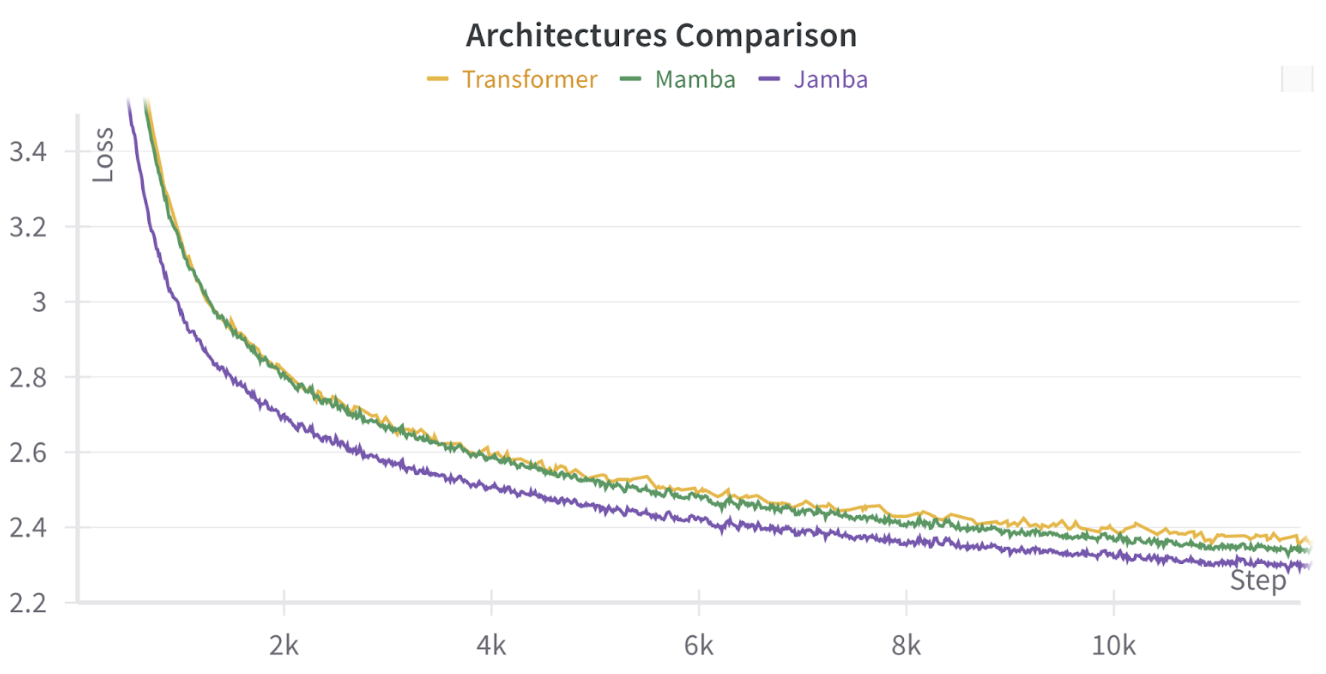}
    \caption{Training loss curves for pure Attention, pure Mamba, and an Attention-Mamba hybrid (no MoE). All models are 7B parameters. The hybrid achieves better loss throughout this training run.}
    \label{fig:loss-attn-mamba-jamba}
\end{figure}

Figure \ref{fig:loss-attn-mamba-jamba} shows the training loss of the three architectures. While the pure Transformer and Mamba models have a similar convergence, the hybrid Jamba (no MoE) has a lower loss throughout this run.

\subsection{Why does the Combination Work?}
\label{sec:mamba-problems}

The pure Mamba model showed fairly good results in most tasks early on, including in general perplexity evaluations. However, it performed substantially worse than the pure Attention model in three common benchmark tasks: IMDB \cite{maas2011learning}, QuAC \cite{choi2018quac}, and NarrativeQA \cite{kocisky2018narrativeqa}. In contrast, the hybrid Attention-Mamba performed similarly to the Attention model on these datasets. 
Table \ref{table:mamba-problems} shows  the results for 1.3B models after 250B tokens. 

\begin{table}[h]
    \centering
    \begin{tabular}{l ccc}
    \toprule 
    & IMDB & QuAC & NarrativeQA \\ 
    \midrule    
    Attention &  84.1 &  27.9 &  45.8 \\ 
    Mamba &  48.8 &  20.2 &  27.7 \\ 
    Attention-Mamba &  90.9 &  26.6 &  43.7 \\ 
    \bottomrule
    \end{tabular}
    \vspace{1pt}
    \caption{Mamba performs poorly on certain datasets, while the Attention-Mamba hybrid performs on par with the Attention model.}
    \label{table:mamba-problems}
\end{table}

Looking into these results further, we found out that the pure Mamba model often does not follow the correct format. For instance, in the IMDB dataset, answer choices are ``Positive'' or ``Negative''. While the Attention model adheres to this format, the pure Mamba model often produces other answers, such as ``Very Good'', ``Very Positive'', ``Funny'', ``Bad'', ``Poor'', and ``3/10''. While these may be considered correct answers, the difficulty of Mamba to adhere to the format suggests a potential problem. Indeed, to perform successful in-context learning, it is important for models to capture the input-output format \cite{min2022rethinking}. The hybrid Attention-Mamba model follows the format successfully, just like the pure Attention model.

We hypothesize that this phenomenon points to a limitation of SSMs -- a potential difficulty in in-context learning (ICL). Indeed, the ability to perform ICL has been linked to the emergence of so-called induction heads in Transformer language models during training, which perform approximate copying operations that are supportive of ICL \cite{olsson2022context}. We conjecture that the lack of an attention mechanism in the pure Mamba model makes it difficult for it to learn in-context. While Mamba may learn to copy and perform simple ICL when explicitly trained to do so (\cite{gu2023mamba,park2024can},  it is not clear if ICL is an emergent capability in SSM as is typical of Transformer models. In contrast, the hybrid Attention–Mamba model does perform successful ICL, even when only 1 out of 8 layers is an Attention one. 

\begin{figure}[h]
    \centering
    \includegraphics[width=\textwidth,trim={3cm 3cm 3cm 3cm},clip]{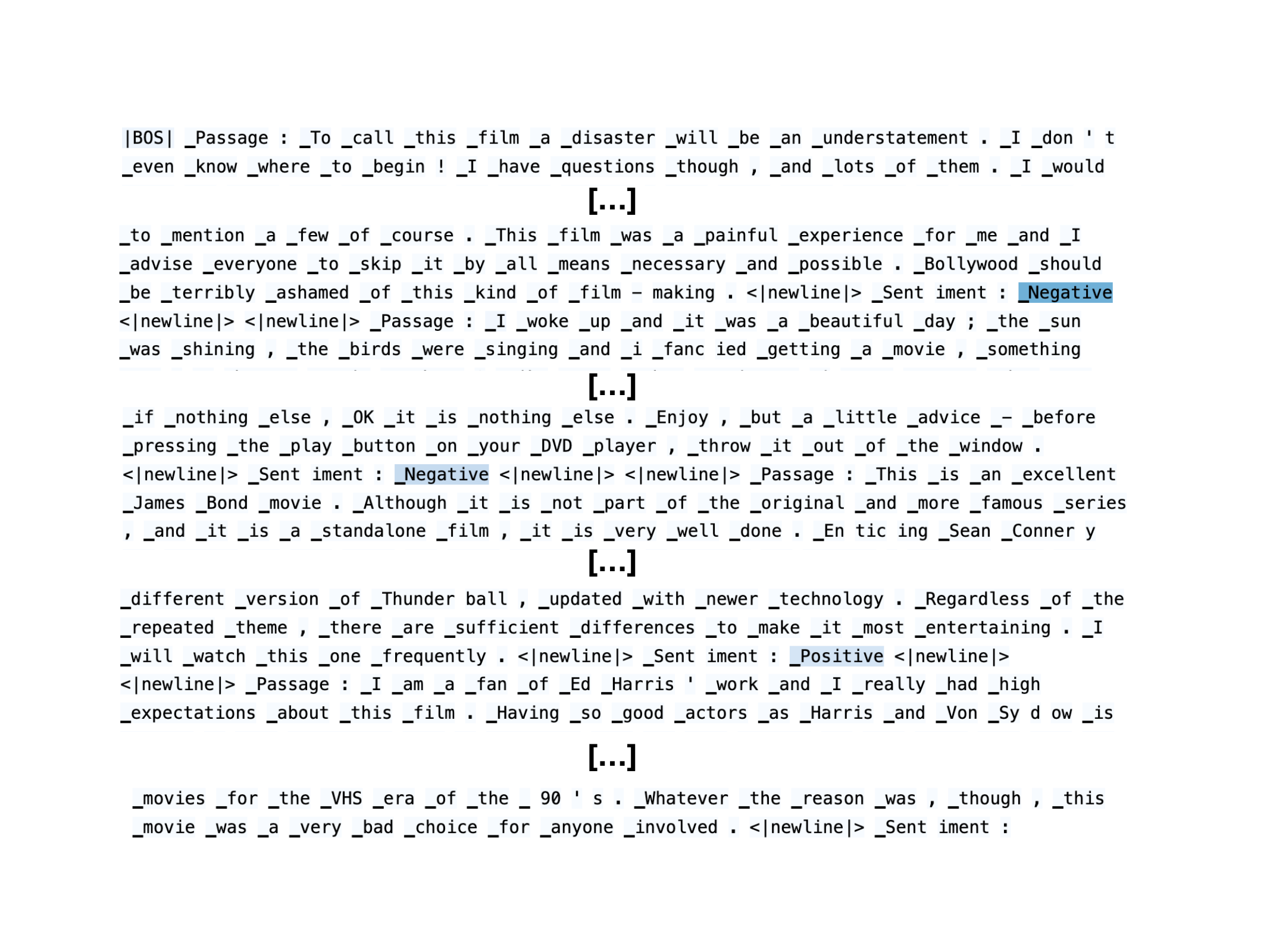}
    \caption{Example induction head (H3, first attention layer) from a hybrid Attention-Mamba model. Highlighted words reflect strong attention from the last token, ``:'',  just before the model is about to predict the label.  We see that the attention is focused on label tokens from the few-shot examples.}
    \label{fig:induction}
\end{figure}

As anecdotal evidence of an emergent induction mechanism, we visualize in Figure \ref{fig:induction} the attention of an example head from a 1.3B Attention-Mamba hybrid model (no MoE), on an IMDB example where the pure Mamba failed and the hybrid succeeded. Clearly, the attention from the last token (``:'') is focused on the labels from the few-shot examples. We have found 12 such heads in our hybrid model, in all three attention layers (which correspond to layers 4, 12, 20 in the model).

Future work can further investigate the emergence of ICL in hybrid models at large scale. Our released checkpoints would hopefully facilitate such investigations. Finally, very recent work has attempted to extract attention-like scores from state-space models like Mamba \cite{ali2024hidden}, which opens another direction to search for induction capabilities in state-space models.

\subsection{The Effect of Mixture-of-Experts (MoE)} \label{sec:moe}

Recent work has shown that MoE improves Transformer language models while keeping compute manageable \cite{jiang2024mixtral}.\footnote{There is also initial evidence that MoE helps Mamba layers, albeit at small model and data scale \cite{pioro2024moe}.} However, it is not clear if MoE integrates well with state-space models at a large scale, and specifically with our hybrid Attention–Mamba architecture. 
Indeed, Table \ref{table:jamba-moe} shows that MoE improves the performance of the hybrid Attention-Mamba architecture at large scale (7B parameters trained on 50B tokens). The MoE variant has $n = 16$ total experts, $K = 2$ experts used at each token, and MoE is applied every $e = 2$ layers, as described in Section \ref{sec:impl-single-gpu}.

\begin{table}[h]
    \centering
    \begin{tabular}{l c c c c c c c}
    \toprule 
         & & \multirow{2}{*}[-5pt]{\makecell[b]{Hella\\Swag}} & \multirow{2}{*}[-5pt]{\makecell[b]{Wino\\Grande}}  & & \multicolumn{3}{c}{log-prob} \\ 
         \cmidrule(lr){6-8}
         & OLLM &  &  & NQ &  C4 &  Books &  Code \\
    \midrule 
    Jamba (no MoE) &  36.6 & 62.5 & 58.8 & 15.4 & -0.547 & -0.658 & -0.340 \\
    Jamba+MoE &  38.1 &  66.0 & 61.2 & 18.9 & -0.534 & -0.645 & -0.326 \\ 
    \bottomrule
    \end{tabular}
    \vspace{1pt}
    \caption{Mixture-of-experts improves the Attention-Mamba hybrid. }
    \label{table:jamba-moe}
\end{table}

\subsection{Stabilizing Mamba at large scale} 

\begin{figure}[h]
    \centering
    \includegraphics[width=.75\textwidth]{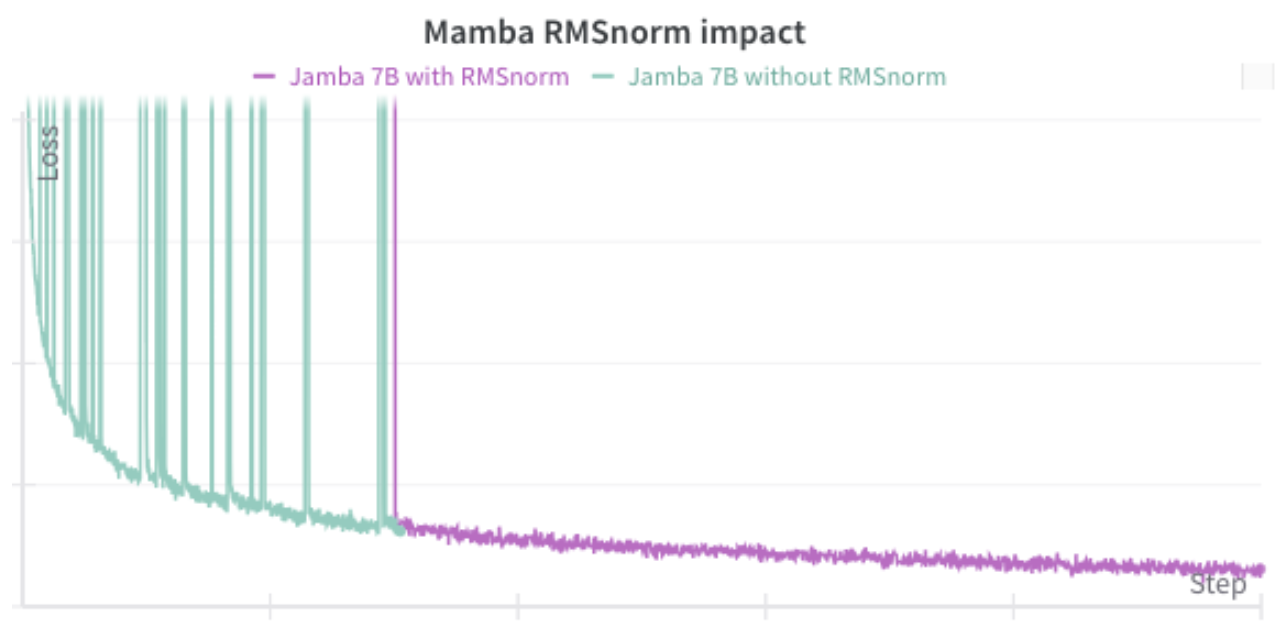}
    \caption{Adding RMSNorm to Mamba layers prevents loss spikes.}
    \label{fig:spike-norm}
\end{figure}

When training Jamba models of up to 1.3B parameters, we observed stable training without special problems. However, when scaling to the largest model released here (7B-based, which has 12B/52B active/total parameters), we encountered large loss spikes. Investigating this revealed that inner parts of the Mamba layers suffer from large activation values, leading to the spikes. We therefore added RMSNorm \cite{zhang2019root} to internal activations. As Figure \ref{fig:spike-norm} shows, this stabilized training and prevented additional loss spikes.

\subsection{Jamba does not Require Explicit Positional Information}

Table \ref{table:position} shows results of the Jamba architecture (with MoE) with no positional information and when applying RoPE \cite{su2024roformer} in the attention layers (1.3B parameter models, 250B tokens). The results are similar, suggesting that explicit positional information may not be required for the hybrid architecture. Presumably, the Mamba layers, which are placed before attention layers, provide implicit position information.\footnote{Some prior evidence suggested that Transformer decoder models do not need positional encodings \cite{haviv2022transformer}. However, all existing large scale models do use some sort of explicit position information.}

\begin{table}[h]
    \centering
    \resizebox{\textwidth}{!}{%
    \begin{tabular}{l c c c c c c c c c c}
    \toprule 
         & & \multirow{2}{*}[-5pt]{\makecell[b]{Hella\\Swag}} & \multirow{2}{*}[-5pt]{\makecell[b]{Wino\\Grande}} & & \multirow{2}{*}[-5pt]{\makecell[b]{Narrative\\QA}} & & & \multicolumn{3}{c}{log-prob} \\ 
         \cmidrule(lr){9-11}
         & OLLM &  &  & ARC-C & & NQ &  BoolQ &  C4 &  Books &  Code \\
    \midrule 
    Jamba &  39.6 & 71.5 & 64.2 & 40.7 & 50.5 & 22.2 & 68.9 & -0.516 & -0.623 & -0.299 \\
    Jamba+RoPE &  40.1  & 71.8 & 65.5 & 40.4 & 46.2 & 22.2 & 67.9 & -0.516 & -0.623 & -0.299 \\ 
    \bottomrule
    \end{tabular}}
    \vspace{1pt}
    \caption{Comparison of Jamba with and without explicit positional information. }
    \label{table:position}
\end{table}

\vspace{-10pt}
\section{Conclusion}

We presented \jamba, a novel architecture which combines Attention and Mamba layers, with MoE modules, and an open implementation of it, reaching state-of-the-art performance and supporting long contexts.
We showed how \jamba provides flexibility for balancing performance and memory requirements, while maintaining a high throughput. We experimented with several design choices such as the ratio of Attention-to-Mamba layers  and discussed some discoveries made during the development process, which will inform future work on hybrid attention--state-space models. To facilitate such research, we plan to release model checkpoints from smaller-scale training runs. The largest model we provide with this release has 12B active and 52B total available parameters, supporting context lengths of up to 256K tokens and fitting in a single 80GB GPU even when processing 140K-token texts.

\bibliographystyle{plain}
\bibliography{refs}

\begin{thebibliography}{10}

\bibitem{ali2024hidden}
Ameen Ali, Itamar Zimerman, and Lior Wolf.
\newblock The hidden attention of mamba models.
\newblock {\em arXiv preprint arXiv:2403.01590}, 2024.

\bibitem{an2023eval}
Chenxin An, Shansan Gong, Ming Zhong, Mukai Li, Jun Zhang, Lingpeng Kong, and Xipeng Qiu.
\newblock {L-Eval}: Instituting standardized evaluation for long context language models.
\newblock {\em arXiv preprint arXiv:2307.11088}, 2023.

\bibitem{bisk2020piqa}
Yonatan Bisk, Rowan Zellers, Jianfeng Gao, Yejin Choi, et~al.
\newblock {PIQA}: Reasoning about physical commonsense in natural language.
\newblock In {\em Proceedings of the AAAI Conference on Artificial Intelligence}, volume~34, pages 7432--7439, 2020.

\bibitem{casanueva2020efficient}
I{\~n}igo Casanueva, Tadas Tem{\v{c}}inas, Daniela Gerz, Matthew Henderson, and Ivan Vuli{\'c}.
\newblock Efficient intent detection with dual sentence encoders.
\newblock In {\em Proceedings of the 2nd Workshop on Natural Language Processing for Conversational AI}, pages 38--45, 2020.

\bibitem{chen2021evaluating}
Mark Chen, Jerry Tworek, Heewoo Jun, Qiming Yuan, Henrique Ponde de~Oliveira Pinto, Jared Kaplan, Harri Edwards, Yuri Burda, Nicholas Joseph, Greg Brockman, et~al.
\newblock Evaluating large language models trained on code.
\newblock {\em arXiv preprint arXiv:2107.03374}, 2021.

\bibitem{choi2018quac}
Eunsol Choi, He~He, Mohit Iyyer, Mark Yatskar, Wen-tau Yih, Yejin Choi, Percy Liang, and Luke Zettlemoyer.
\newblock {QuAC}: Question answering in context.
\newblock In {\em Proceedings of the 2018 Conference on Empirical Methods in Natural Language Processing}, pages 2174--2184, 2018.

\bibitem{chowdhery2023palm}
Aakanksha Chowdhery, Sharan Narang, Jacob Devlin, Maarten Bosma, Gaurav Mishra, Adam Roberts, Paul Barham, Hyung~Won Chung, Charles Sutton, Sebastian Gehrmann, et~al.
\newblock Palm: Scaling language modeling with pathways.
\newblock {\em Journal of Machine Learning Research}, 24(240):1--113, 2023.

\bibitem{clark2022unified}
Aidan Clark, Diego de~Las~Casas, Aurelia Guy, Arthur Mensch, Michela Paganini, Jordan Hoffmann, Bogdan Damoc, Blake Hechtman, Trevor Cai, Sebastian Borgeaud, et~al.
\newblock Unified scaling laws for routed language models.
\newblock In {\em International conference on machine learning}, pages 4057--4086. PMLR, 2022.

\bibitem{clark2019boolq}
Christopher Clark, Kenton Lee, Ming-Wei Chang, Tom Kwiatkowski, Michael Collins, and Kristina Toutanova.
\newblock {BoolQ}: Exploring the surprising difficulty of natural yes/no questions.
\newblock In {\em Proceedings of the 2019 Conference of the North American Chapter of the Association for Computational Linguistics: Human Language Technologies, Volume 1 (Long and Short Papers)}, pages 2924--2936, 2019.

\bibitem{clark2018think}
Peter Clark, Isaac Cowhey, Oren Etzioni, Tushar Khot, Ashish Sabharwal, Carissa Schoenick, and Oyvind Tafjord.
\newblock Think you have solved question answering? try {ARC}, the {AI2} reasoning challenge.
\newblock {\em arXiv preprint arXiv:1803.05457}, 2018.

\bibitem{cobbe2021training}
Karl Cobbe, Vineet Kosaraju, Mohammad Bavarian, Mark Chen, Heewoo Jun, Lukasz Kaiser, Matthias Plappert, Jerry Tworek, Jacob Hilton, Reiichiro Nakano, et~al.
\newblock Training verifiers to solve math word problems.
\newblock {\em arXiv preprint arXiv:2110.14168}, 2021.

\bibitem{ollm}
Hugging Face.
\newblock Open {LLM} leaderboard.
\newblock \url{https://huggingface.co/spaces/HuggingFaceH4/open_llm_leaderboard}, 2024.

\bibitem{fathullah23_interspeech}
Yassir Fathullah, Chunyang Wu, Yuan Shangguan, Junteng Jia, Wenhan Xiong, Jay Mahadeokar, Chunxi Liu, Yangyang Shi, Ozlem Kalinli, Mike Seltzer, and Mark J.~F. Gales.
\newblock Multi-head state space model for speech recognition.
\newblock In {\em Proceedings of INTERSPEECH 2023}, pages 241--245, 2023.

\bibitem{fedus2022switch}
William Fedus, Barret Zoph, and Noam Shazeer.
\newblock Switch transformers: Scaling to trillion parameter models with simple and efficient sparsity.
\newblock {\em Journal of Machine Learning Research}, 23(120):1--39, 2022.

\bibitem{fu2022hungry}
Daniel~Y Fu, Tri Dao, Khaled~Kamal Saab, Armin~W Thomas, Atri Rudra, and Christopher Re.
\newblock Hungry hungry hippos: Towards language modeling with state space models.
\newblock In {\em The Eleventh International Conference on Learning Representations}, 2022.

\bibitem{gage1994new}
Philip Gage.
\newblock A new algorithm for data compression.
\newblock {\em The C Users Journal}, 12(2):23--38, 1994.

\bibitem{gu2023mamba}
Albert Gu and Tri Dao.
\newblock Mamba: Linear-time sequence modeling with selective state spaces.
\newblock {\em arXiv preprint arXiv:2312.00752}, 2023.

\bibitem{gu2021efficiently}
Albert Gu, Karan Goel, and Christopher Re.
\newblock Efficiently modeling long sequences with structured state spaces.
\newblock In {\em International Conference on Learning Representations}, 2021.

\bibitem{gu2021combining}
Albert Gu, Isys Johnson, Karan Goel, Khaled Saab, Tri Dao, Atri Rudra, and Christopher R{\'e}.
\newblock Combining recurrent, convolutional, and continuous-time models with linear state space layers.
\newblock {\em Advances in neural information processing systems}, 34:572--585, 2021.

\bibitem{haviv2022transformer}
Adi Haviv, Ori Ram, Ofir Press, Peter Izsak, and Omer Levy.
\newblock Transformer language models without positional encodings still learn positional information.
\newblock In {\em Findings of the Association for Computational Linguistics: EMNLP 2022}, pages 1382--1390, 2022.

\bibitem{hendrycks2020measuring}
Dan Hendrycks, Collin Burns, Steven Basart, Andy Zou, Mantas Mazeika, Dawn Song, and Jacob Steinhardt.
\newblock Measuring massive multitask language understanding.
\newblock In {\em International Conference on Learning Representations}, 2020.

\bibitem{hendrycks2021cuad}
Dan Hendrycks, Collin Burns, Anya Chen, and Spencer Ball.
\newblock {CUAD}: An expert-annotated {NLP} dataset for legal contract review.
\newblock In {\em Thirty-fifth Conference on Neural Information Processing Systems Datasets and Benchmarks Track (Round 1)}, 2021.

\bibitem{jiang2023mistral}
Albert~Q Jiang, Alexandre Sablayrolles, Arthur Mensch, Chris Bamford, Devendra~Singh Chaplot, Diego de~las Casas, Florian Bressand, Gianna Lengyel, Guillaume Lample, Lucile Saulnier, et~al.
\newblock Mistral 7b.
\newblock {\em arXiv preprint arXiv:2310.06825}, 2023.

\bibitem{jiang2024mixtral}
Albert~Q Jiang, Alexandre Sablayrolles, Antoine Roux, Arthur Mensch, Blanche Savary, Chris Bamford, Devendra~Singh Chaplot, Diego de~las Casas, Emma~Bou Hanna, Florian Bressand, et~al.
\newblock Mixtral of experts.
\newblock {\em arXiv preprint arXiv:2401.04088}, 2024.

\bibitem{kamradt2023}
Greg Kamradt.
\newblock Needle in a haystack - pressure testing llms.
\newblock \url{https://github.com/gkamradt/LLMTest_NeedleInAHaystack/}, 2023.

\bibitem{kocisky2018narrativeqa}
Tomas Kocisky, Jonathan Schwarz, Phil Blunsom, Chris Dyer, Karl~Moritz Hermann, Gabor Melis, and Edward Grefenstette.
\newblock The {NarrativeQA} reading comprehension challenge.
\newblock {\em Transactions of the Association for Computational Linguistics}, 6:317--328, 2018.

\bibitem{kwiatkowski2019natural}
Tom Kwiatkowski, Jennimaria Palomaki, Olivia Redfield, Michael Collins, Ankur Parikh, Chris Alberti, Danielle Epstein, Illia Polosukhin, Jacob Devlin, Kenton Lee, et~al.
\newblock Natural questions: a benchmark for question answering research.
\newblock {\em Transactions of the Association for Computational Linguistics}, 7:452--466, 2019.

\bibitem{larson2019evaluation}
Stefan Larson, Anish Mahendran, Joseph~J Peper, Christopher Clarke, Andrew Lee, Parker Hill, Jonathan~K Kummerfeld, Kevin Leach, Michael~A Laurenzano, Lingjia Tang, et~al.
\newblock An evaluation dataset for intent classification and out-of-scope prediction.
\newblock In {\em Proceedings of the 2019 Conference on Empirical Methods in Natural Language Processing and the 9th International Joint Conference on Natural Language Processing (EMNLP-IJCNLP)}, pages 1311--1316, 2019.

\bibitem{li2002learning}
Xin Li and Dan Roth.
\newblock Learning question classifiers.
\newblock In {\em COLING 2002: The 19th International Conference on Computational Linguistics}, 2002.

\bibitem{lin-etal-2022-truthfulqa}
Stephanie Lin, Jacob Hilton, and Owain Evans.
\newblock {T}ruthful{QA}: Measuring how models mimic human falsehoods.
\newblock In {\em Proceedings of the 60th Annual Meeting of the Association for Computational Linguistics (Volume 1: Long Papers)}, pages 3214--3252, Dublin, Ireland, May 2022. Association for Computational Linguistics.

\bibitem{liu2021benchmarking}
Xingkun Liu, Arash Eshghi, Pawel Swietojanski, and Verena Rieser.
\newblock Benchmarking natural language understanding services for building conversational agents.
\newblock In {\em Increasing Naturalness and Flexibility in Spoken Dialogue Interaction: 10th International Workshop on Spoken Dialogue Systems}, pages 165--183. Springer, 2021.

\bibitem{maas2011learning}
Andrew Maas, Raymond~E Daly, Peter~T Pham, Dan Huang, Andrew~Y Ng, and Christopher Potts.
\newblock Learning word vectors for sentiment analysis.
\newblock In {\em Proceedings of the 49th annual meeting of the association for computational linguistics: Human language technologies}, pages 142--150, 2011.

\bibitem{mielke2021between}
Sabrina~J Mielke, Zaid Alyafeai, Elizabeth Salesky, Colin Raffel, Manan Dey, Matthias Gall{\'e}, Arun Raja, Chenglei Si, Wilson~Y Lee, Beno{\^\i}t Sagot, et~al.
\newblock Between words and characters: A brief history of open-vocabulary modeling and tokenization in {NLP}.
\newblock {\em arXiv preprint arXiv:2112.10508}, 2021.

\bibitem{min2022rethinking}
Sewon Min, Xinxi Lyu, Ari Holtzman, Mikel Artetxe, Mike Lewis, Hannaneh Hajishirzi, and Luke Zettlemoyer.
\newblock Rethinking the role of demonstrations: What makes in-context learning work?
\newblock In {\em Proceedings of the 2022 Conference on Empirical Methods in Natural Language Processing}, pages 11048--11064, 2022.

\bibitem{olsson2022context}
Catherine Olsson, Nelson Elhage, Neel Nanda, Nicholas Joseph, Nova DasSarma, Tom Henighan, Ben Mann, Amanda Askell, Yuntao Bai, Anna Chen, et~al.
\newblock In-context learning and induction heads.
\newblock {\em arXiv preprint arXiv:2209.11895}, 2022.

\bibitem{park2024can}
Jongho Park, Jaeseung Park, Zheyang Xiong, Nayoung Lee, Jaewoong Cho, Samet Oymak, Kangwook Lee, and Dimitris Papailiopoulos.
\newblock Can mamba learn how to learn? a comparative study on in-context learning tasks.
\newblock {\em arXiv preprint arXiv:2402.04248}, 2024.

\bibitem{pilault2023block}
Jonathan Pilault, Mahan Fathi, Orhan Firat, Christopher Pal, Pierre-Luc Bacon, and Ross Goroshin.
\newblock Block-state transformers.
\newblock In {\em Thirty-seventh Conference on Neural Information Processing Systems}, 2023.

\bibitem{pioro2024moe}
Maciej Pi{\'o}ro, Kamil Ciebiera, Krystian Kr{\'o}l, Jan Ludziejewski, and Sebastian Jaszczur.
\newblock {MoE-Mamba}: Efficient selective state space models with mixture of experts.
\newblock {\em arXiv preprint arXiv:2401.04081}, 2024.

\bibitem{poli2023hyena}
Michael Poli, Stefano Massaroli, Eric Nguyen, Daniel~Y Fu, Tri Dao, Stephen Baccus, Yoshua Bengio, Stefano Ermon, and Christopher R{\'e}.
\newblock Hyena hierarchy: Towards larger convolutional language models.
\newblock In {\em International Conference on Machine Learning}, pages 28043--28078. PMLR, 2023.

\bibitem{stripedhyena}
Michael Poli, Jue Wang, Stefano Massaroli, Jeffrey Quesnelle, Ryan Carlow, Eric Nguyen, and Armin Thomas.
\newblock {StripedHyena: Moving Beyond Transformers with Hybrid Signal Processing Models}.
\newblock \url{ https://github.com/togethercomputer/stripedhyena }, 2023.

\bibitem{ratner2023parallel}
Nir Ratner, Yoav Levine, Yonatan Belinkov, Ori Ram, Inbal Magar, Omri Abend, Ehud Karpas, Amnon Shashua, Kevin Leyton-Brown, and Yoav Shoham.
\newblock Parallel context windows for large language models.
\newblock In {\em Proceedings of the 61st Annual Meeting of the Association for Computational Linguistics (Volume 1: Long Papers)}, pages 6383--6402, 2023.

\bibitem{sakaguchi2020winogrande}
Keisuke Sakaguchi, Ronan Le~Bras, Chandra Bhagavatula, and Yejin Choi.
\newblock {WinoGrande}: An adversarial winograd schema challenge at scale.
\newblock In {\em Proceedings of the AAAI Conference on Artificial Intelligence}, volume~34, pages 8732--8740, 2020.

\bibitem{saon2023diagonal}
George Saon, Ankit Gupta, and Xiaodong Cui.
\newblock Diagonal state space augmented transformers for speech recognition.
\newblock In {\em ICASSP 2023-2023 IEEE International Conference on Acoustics, Speech and Signal Processing (ICASSP)}, pages 1--5. IEEE, 2023.

\bibitem{sennrich2016neural}
Rico Sennrich, Barry Haddow, and Alexandra Birch.
\newblock Neural machine translation of rare words with subword units.
\newblock In {\em Proceedings of the 54th Annual Meeting of the Association for Computational Linguistics (Volume 1: Long Papers)}, pages 1715--1725, 2016.

\bibitem{shazeer2020glu}
Noam Shazeer.
\newblock Glu variants improve transformer.
\newblock {\em arXiv preprint arXiv:2002.05202}, 2020.

\bibitem{shazeer2016outrageously}
Noam Shazeer, Azalia Mirhoseini, Krzysztof Maziarz, Andy Davis, Quoc Le, Geoffrey Hinton, and Jeff Dean.
\newblock Outrageously large neural networks: The sparsely-gated mixture-of-experts layer.
\newblock In {\em International Conference on Learning Representations}, 2017.

\bibitem{su2024roformer}
Jianlin Su, Murtadha Ahmed, Yu~Lu, Shengfeng Pan, Wen Bo, and Yunfeng Liu.
\newblock Roformer: Enhanced transformer with rotary position embedding.
\newblock {\em Neurocomputing}, 568:127063, 2024.

\bibitem{suzgun2023challenging}
Mirac Suzgun, Nathan Scales, Nathanael Sch{\"a}rli, Sebastian Gehrmann, Yi~Tay, Hyung~Won Chung, Aakanksha Chowdhery, Quoc Le, Ed~Chi, Denny Zhou, et~al.
\newblock Challenging {BIG-Bench} tasks and whether chain-of-thought can solve them.
\newblock In {\em Findings of the Association for Computational Linguistics: ACL 2023}, pages 13003--13051, 2023.

\bibitem{team2024gemma}
Gemma Team, Thomas Mesnard, Cassidy Hardin, Robert Dadashi, Surya Bhupatiraju, Shreya Pathak, Laurent Sifre, Morgane Rivi{\`e}re, Mihir~Sanjay Kale, Juliette Love, et~al.
\newblock Gemma: Open models based on gemini research and technology.
\newblock {\em arXiv preprint arXiv:2403.08295}, 2024.

\bibitem{touvron2023llama}
Hugo Touvron, Louis Martin, Kevin Stone, Peter Albert, Amjad Almahairi, Yasmine Babaei, Nikolay Bashlykov, Soumya Batra, Prajjwal Bhargava, Shruti Bhosale, et~al.
\newblock Llama 2: Open foundation and fine-tuned chat models.
\newblock {\em arXiv preprint arXiv:2307.09288}, 2023.

\bibitem{vaswani2017attention}
Ashish Vaswani, Noam Shazeer, Niki Parmar, Jakob Uszkoreit, Llion Jones, Aidan~N Gomez, {\L}ukasz Kaiser, and Illia Polosukhin.
\newblock Attention is all you need.
\newblock {\em Advances in neural information processing systems}, 30, 2017.

\bibitem{zellers2019hellaswag}
Rowan Zellers, Ari Holtzman, Yonatan Bisk, Ali Farhadi, and Yejin Choi.
\newblock {HellaSwag}: Can a machine really finish your sentence?
\newblock In {\em Proceedings of the 57th Annual Meeting of the Association for Computational Linguistics}, pages 4791--4800, 2019.

\bibitem{zhang2019root}
Biao Zhang and Rico Sennrich.
\newblock Root mean square layer normalization.
\newblock {\em Advances in Neural Information Processing Systems}, 32, 2019.

\bibitem{zoph2022st}
Barret Zoph, Irwan Bello, Sameer Kumar, Nan Du, Yanping Huang, Jeff Dean, Noam Shazeer, and William Fedus.
\newblock {ST-MoE}: Designing stable and transferable sparse expert models.
\newblock {\em arXiv preprint arXiv:2202.08906}, 2022.

\bibitem{zuo2022efficient}
Simiao Zuo, Xiaodong Liu, Jian Jiao, Denis Charles, Eren Manavoglu, Tuo Zhao, and Jianfeng Gao.
\newblock Efficient long sequence modeling via state space augmented transformer.
\newblock {\em arXiv preprint arXiv:2212.08136}, 2022.

\end{thebibliography}

\end{document}